\documentclass{ws-procs9x6}            % default, citations in superscript

\usepackage{algorithm}
\usepackage{algorithmic}
\usepackage{subfigure}
\usepackage{graphicx}
\begin{document}
\title{Metric learning by Similarity Network for Deep Semi-Supervised Learning}

\author{Sanyou Wu, Xingdong Feng, Fan Zhou\footnote{Corresponding author.}}

\address{School of Statistics and Management,\\
Shanghai University of Finance and Economics,\\
Shanghai, 200433, China\\
$^*$E-mail: zhoufan@mail.shufe.edu.cn}

\begin{abstract}
Deep semi-supervised learning has been widely implemented in the real-world due to the rapid development of deep learning. Recently, attention has shifted to the approaches such as Mean-Teacher to penalize the inconsistency between two perturbed input sets. Although these methods may achieve positive results, they ignore the relationship information between data instances. To solve this problem, we propose a novel method named Metric Learning by Similarity Network (MLSN), which aims to learn a distance metric adaptively on different domains. By co-training with the classification network, similarity network can learn more information about pairwise relationships and performs better on some empirical tasks than state-of-art methods.  
%and make representations of feature space more separable. Empirically, it has more effective performance than the Mean-Teacher algorithm on benchmark datasets, SVHN and CIFAR-10 respectively.
\end{abstract}


\bodymatter

\section{Introduction}\label{aba:sec1}
The success of deep learning in many complicated tasks, such as image classification, speech recognition, and machine translation usually relies on a sufficient number of labeled training samples. However, acquiring manually labeled data is expensive and time-consuming in practice. In contrast, the training sample with only a small portion of labeled instances is more easily to achieve most of time. In this case, Semi-Supervised Learning (SSL) will be used to utilize the unlabeled data by leveraging the labeled ones. 

More formally, let $X = (x_1,x_2,\ldots,x_n)$ be a set of $n$ samples. $x_i\in \mathcal{X}$ for each $i$, where $\mathcal{X}$ is a sample space. Only the first $l$ examples are labeled with $y$ and their label set is denoted by $Y_L=\{y_1,y_2,\ldots,y_l\}$. We let $U=\{x_{l+1},x_{l+2},\ldots,x_n\}$ be the set of unlabeled data consisted of remaining unlabeled examples. The goal of SSL is to learn a model $f(x;\theta)=p(y|x,\theta)$ which utilizes both labeled and unlabeled data, and assign a prediction to each of the $|U|$ unlabeled instances. 

One basic assumption of SSL \citep{zhu2009introduction,mey2019improvability} is that, instances sharing similar representations are more likely to be assigned the same labels. However, most existing methods \citep{dai2017good,chongxuan2017triple,laine2016temporal,tarvainen2017mean,miyato2018virtual,belkin2006manifold,iscen2019label} ignore the pairwise relationship within training samples, including both labeled and unlabeled ones. 
To tackle this issue, we propose a method called Metric Learning by Similarity Network (MLSN) based on Mean-Teacher method \cite{tarvainen2017mean} to efficiently learn the pairwise information by incorporating a similarity network. Our contributions are summarized as follows: 
\begin{itemlist}[(3)]
	\item We propsed an approach to learn a similarity metric adaptively for data in different domains. And our method can capture the relationship information between data instances.
	\item We use neural networks to learn the similarity between instances and make different classes more separable in the feature space.
	\item Our approach can be easily implemented into other existing neural networks for SSL.
	%\item Based on MT framework, our method performs better than other state-of-art algorithms on the benchmark datasets SVHN and CIFAR-10.
\end{itemlist}

\section{Our approach}
%In graph-based algorithms, the quality of the adjacency matrix directly determines the performance of the model. Traditionally, a pre-defined distance is often used to measure the distance between two instances $(x_i,x_j)$, e.g., Euclidean distance and Gaussian distance. For image data, it is too difficult to define the similarity $s(x_i,x_j)$ between $x_i$ and $x_j$. The key idea of this paper is utilizing neural networks to learn the similarity between images.\par
%Another advantage of our algorithm is that it can make the feature space of pictures more separable. This intuition is consistent with a series of works of literature on face recognition \cite{schroff2015facenet,liu2017sphereface,wang2018cosface,deng2019arcface}. Under clustering assumption of SSL, the more separable feature space always means better performance of the model.
\subsection{Learning the metric by similarity network} %3.1
\begin{figure}
	\begin{center}
		\includegraphics[width=3.5in]{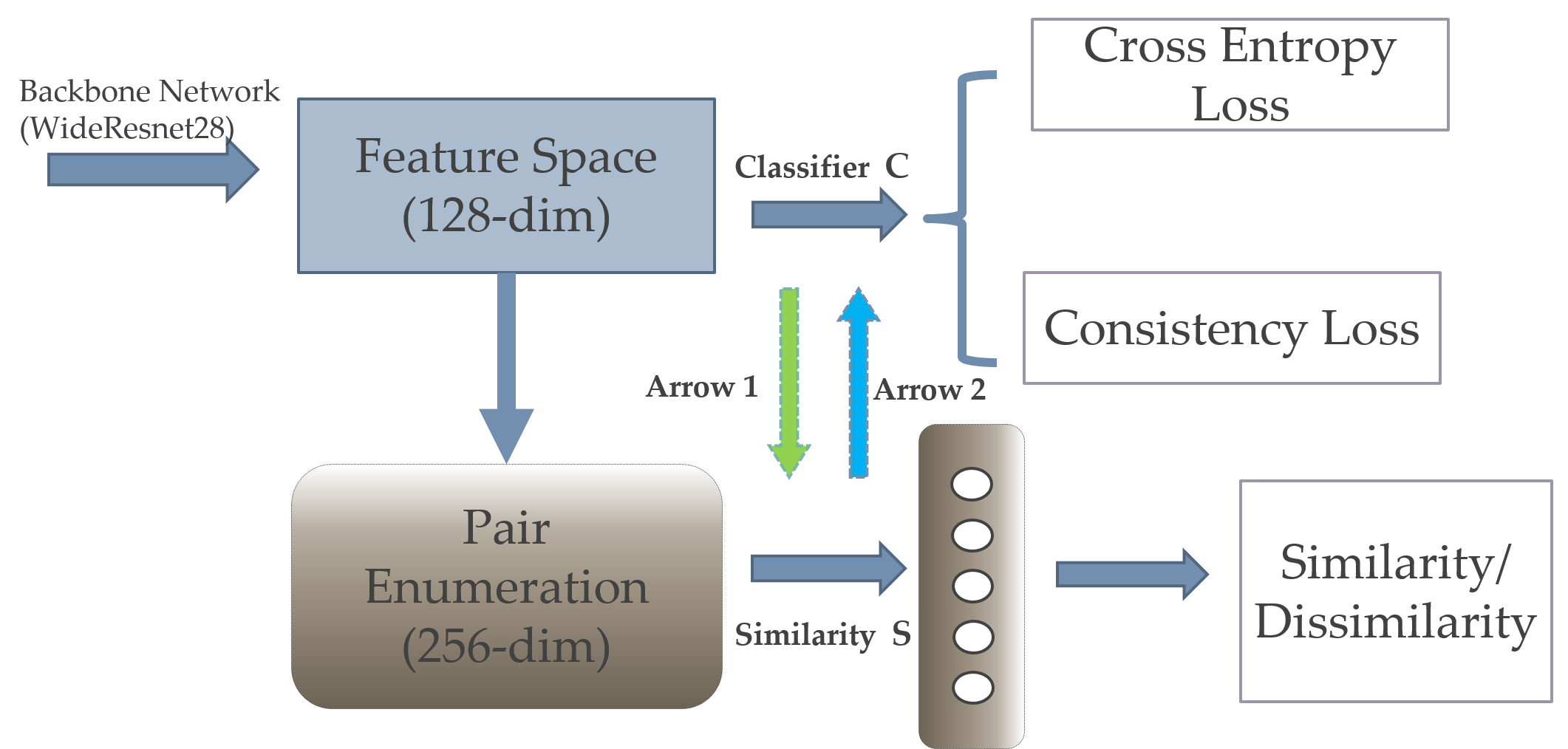}
	\end{center}
	\caption{The network architecture of our proposed MLSN, in which we append a similarity network branch to learn the metric based on the Mean-Teacher framework.}
	\label{aba:fig1}
\end{figure}

%According to Universal Approximation Theorem \cite{hornik1991approximation,cybenko1989approximation} (UAT), we create a similar network to approximate the similarity function. 
The whole network architecture of our model is illustrated in Figure \ref{aba:fig1}. 
%In general, a deep neural network $f$ can be decomposed as $f=C \circ h$, where $h:\mathcal{X} \rightarrow R^p$ is the mapping from the input space to the feature space and $C:R^p\rightarrow [0,1]^K$ is the output layer usually parameterized by a fully-connected layer with softmax. 
A Wide-ResNet28 \cite{zagoruyko2016wide} is used as feature extractor $h$, where $h:\mathcal{X} \rightarrow R^p$ is the mapping from the input space to the feature space, and convert input data to 128-dim vectors, denote this as feature space. 
After getting the features, we apply a two-branches network. One is a classification network, which employs classification loss (e.g., cross-entropy) and consistency loss (e.g., squared loss) in SSL. The other branch is a similarity network used to approximate the similarity function, which is denoted by $S$, which determines the probability that two samples $(x_i,x_j)$ are similar. The similarity loss $L_S$ is defined as follows, 
\begin{equation}
L_S= \sum_{i,j\in L}d(S(x_i,x_j ),y_{simi})
\end{equation} 
Where $d(.,.)$ is a pre-defined distance function, $y_{simi}$ is the label of $S$, $S(x_i,x_j)$ is the probability prediction of an arbitrary sample pair $(x_i,x_j )$.
\begin{equation}
y_{simi} = \left\{\begin{matrix}
1, & \text{if}  &  {y_i} = {y_j}\\
0, & \text{else} &
\end{matrix}\right.
\end{equation}

We introduce a simple example to show how converting weak labels into pseudo labels is more efficient than traditional approaches, for example, the self-training method. 

% Unnumbered appendix sections can be obtained using \verb|\section*|.
%\begin{figure}
%	\begin{center}
%		\includegraphics[width=4.5in]{simi-label}
%	\end{center}
%	\caption{(a,b,c) are the visualization of feature space using t-SNE . (a) 1000 lables (b) unlabeled samples with self-training pseudo-label with high confidence (c) unlabeled samples with similarity pseudo-label with high confidence. }
%	\label{aba:fig3}
%\end{figure}
\begin{figure}
	\begin{center}
		\subfigure[labeled]{
			\label{sub1}
			\includegraphics[width=0.3\textwidth]{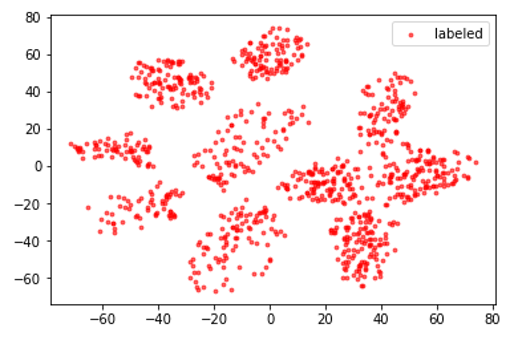}
		}
		\subfigure[self-training]{
			\label{sub2}
			\includegraphics[width=0.3\textwidth]{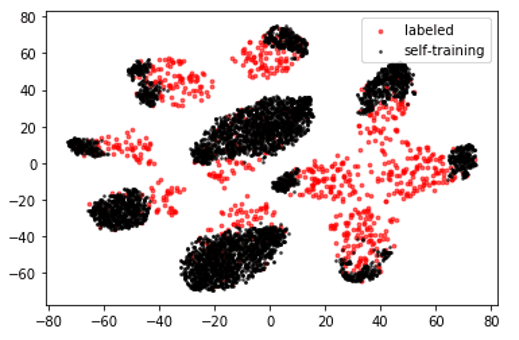}
		}
		\subfigure[similairty]{
			\includegraphics[width=0.3\textwidth]{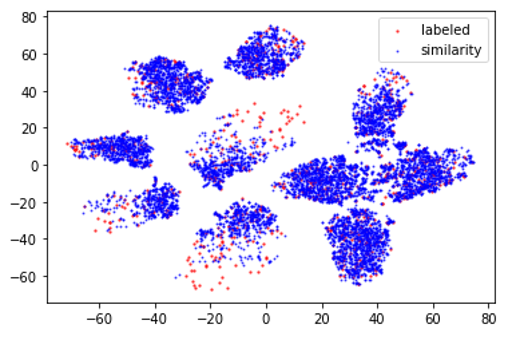}
		}
	\end{center}
	\caption{(a,b,c) are the visualization of feature space using t-SNE . (a) 1000 lables (b) unlabeled samples with self-training pseudo-label with high confidence (c) unlabeled samples with similarity pseudo-label with high confidence. }
	\label{aba:fig3}
\end{figure}

On the CIFAR-10 dataset, we train a model using 1000 labeled samples and use self-training and similarity pseudo-label to improve the classifier, respectively. As shown in Figure \ref{aba:fig3}, the feature distribution of unlabeled instances learned by similarity pseudo-labels (blue points) are more significantly separated from that of the labeled data (red points) compared to self-training (balck points). This tells that the main advantage of similarity pseudo-labels is to expand the feature space of labeled data and offer more information for classifier during training.

%In the \fref{aba:fig3}, the features correspond to labeled data extacted by the model just visulized for comparison, and its shown in read on (a). Unlabeled samples are shown in black and blue respectively on (b) and (c).

%%%--3.2
\subsection{Co-training with classification network and similarity network}

Inspired by GAN, we employ a co-training approach to jointly learn the classification network $C$ and the similarity network $S$. We formalize the way to generate pseudo labels for network $S$ and network $C$, respectively.
%We consider two ways to generate pseudo labels for network $C$ by using network $S$. %where the outputs of two networks serve as their inputs, respectively. 

\textbf{Label for similarity network}. Different from the self-training method, the pseudo-labels obtained by the classifier are not directly incorporated into the training sample. Instead, they are converted into weak labels and serve as the input of the similarity network $S$. For each pair of unlabeled input features $(x_i,x_j)$, we can generate the corresponding pseudo labels $\hat{y}_i,\hat{y}_j$ using the classification network $C$. If $\hat{y}_i= \hat{y}_j$, object $i$ and $j$ are considered to be similar. Thus, the similarity label of pair $(x_i,x_j)$ in the similarity network equals to $1$ in this case, and equals to $0$ otherwise. The modified definition of $y_{simi}$ by replacing $y_i$ by its prediction $\hat y_i$ is defined as follows,
%are no longer as the input to classifier repeatedly. Instead, it is first to convert into a weak label and input into the similar network $S$. For unlabeled data $(x_i,x_j)$ in labeled and unlabeled set, we can get the pseudo label $\hat{y}_i,\hat{y}_j$ by Classifier network. If $\hat{y}_i= \hat{y}_j$, we think the instances $x_i$ and $x_j$ are similar. And its label for similarity network (similarity label) equals 1. On the contrary, the similarity label equals 0 when $\hat{y}_i \neq \hat{y}_j$. And, the label of Similarity label also remarked as $y_{simi}$.
\begin{equation}
	y_{simi} = \left\{\begin{matrix}
	1,& \text{if} & \hat{y}_i = \hat{y}_j\\
	0,& \text{else} &
	\end{matrix}\right.
\end{equation}

\textbf{Label for classification network}. Basically, each input pair $x_i,x_j$ to the similarity network itself is a certain kind of weak label. Therefore, we can generate another kind of label which can be used by network $C$, and let the pseudo label obtained in this way be $y_{sc}$.
%is a kind of weak label. Therefore, the core is to  convert the weak information to the normal label needed by the classifier. We mark labeled data mini-batch is $BL$, and unlabeled data mini-batch is $BU$.
\begin{itemize}
    \item For labeled data in a batch, we randomly choose a sample as the center for every class. Such as in dataset CIFAR-10 and SVHN, we can obtain $K$ class centers $(c_1,c_2,\ldots,c_K)$ in every batch.
    \item Obtain $(x_i,c_j)$ with unlabeled data $x_i \in BU$ and centers of each pre-defined classes $c_j \in (c_1,c_2,\ldots,c_K)$, which is fed into similarity network to get the similarity between instance $i$ and all the $K$ classes, denoted by $p_1,p_2,\ldots,p_K$. $BU$ here denotes a mini-batch sampled from the unlabeled set. 
    \item Treat $p_1,p_2,\ldots,p_K$ as the soft-label set for unlabeled instance $i$, where a larger $p_i$ reprents a higher similarity. 
    %Intuitively, the larger $p_j,j=1,2,…,10$ is, the more similar $x_i$ is to class j and it is consistent with the one-hot result of Classifier network.
\end{itemize}

\subsection{Computation Complexity}
Our overall objective function is the sum of the classification network loss and the similarity network loss. For the classification network, we have two components. The first one is the standard cross-entropy loss on the labeled data, and the second is the consistency cost function $J(\theta)$ used to reduce prediction variance. Algorithm \ref{alg:1} illustrates the detailed training process and $BL$ denotes the mini-batch sampled from the labeled set. 
%. For Similarity network, the loss function $L_S$ is mentioned above. Alg.1 presents the pseudo-code.\par

In each iteration, we have two mini-batch of size $n$ from labeled and unlabeled sets, respectively. Since the similarity network uses instance pairs $(x_i,x_j)$'s of size $n^2$ and the co-training step consider the instance-center pairs of size $n*K$, the overall computational complexity is $n^2+nK$.

To reduce the computation complexity, we use a stochastic sample method to sample $m$ ($m \ll n$) instance pairs to train $S$ and use Focal Loss \cite{lin2017focal} for similarity loss $L_S$, to achieve the balance between similarity and dissimilarity cases. Empirically, the size $n$ falls into range $[64,100]$, and the overall time cost does not vary a lot across different epochs.

\begin{algorithm}
	\caption{Mini-batch training of MLSN for SSL}
	\label{alg:1}
	\begin{algorithmic}[1]
		\REQUIRE $f_\theta(x)$    = neural networks with trainable parameters $\theta$\\
		\REQUIRE $D_L(x,y)$     = set of labeled data\\
		\REQUIRE $D_U(x)$          = set of unlabeled data \\
		\REQUIRE $\lambda_1(t)$    = weight ramp-up function for Consistency loss \\
		\REQUIRE $\lambda_2(t)$  = weight ramp-up function for Similarity loss \\
		\REQUIRE $\lambda_3(t)$    = weight ramp-up function for Co-training loss \\
		%		\STATE {\bfseries Input:} $x_i, y_i$\\
		%		$w(t) = $ unsupervised weight ramp-up function\\
		%		$f_\theta(x) = $ neural network with trainable parameters $\theta$
		\FOR {$t$ in [1, numepochs]}
		\FOR {each minibatch $BL,BU$}
		\STATE {$f_\theta(x) \leftarrow$ evaluate classifier outputs \\
		$L_C = CrossEntropy(\{f_\theta(x_i),y_i\}_{i=1}^{BL})$ \quad - Supervised loss}

		 \STATE{$\tilde{f}_{\theta}(x) \leftarrow$ calculated by the teacher model \\
		 $L_T = d(\tilde{f}_{\theta}(x),f_\theta(x))$ \quad \quad \quad \qquad - Consistency loss}
		 \smallskip
		 \STATE{sampling pairs set $S_1 = (x_i,x_j) \in BL,BU$  \\
		 $s(x_i,x_j) \leftarrow$ evaluated similarity network outputs \\
	 	 $L_S = d(s(x_i,x_j),y_{simi})$ - Similarity loss (e.g., Focal loss)}
		\STATE {similarity-label set $S_2 = (x_i,y_{sc}),x_i \in BU$ \\
		$L_{SC} = CrossEntropy(f_\theta(x),y_{sc})$ - Co-training loss }
		\STATE {$loss_{total} = L_C + \lambda_1(t)*L_T + \lambda_2(t)*L_S + \lambda_3(t)*L_{SC}$}
		\ENDFOR	
		\ENDFOR
		\STATE{return $\theta$}
	\end{algorithmic}
\end{algorithm}
%\label{sec:alg}

\section{Experiments}
\subsection{Datasets}
We follow the common practice in semi-supervised learning literature and conduct experiments using the Street View House Numbers \cite{netzer2011reading} (SVHN) and CIFAR-10 datasets. The SVHN contains 32x32 pixel RGB images of real-world house numbers and belonging to 10 classes. In SVHN, there are 73257 training sample and 26032 test samples.\par
The CIFAR-10 dataset also consists of 32x32 images from 10 classes. There are all 60,000 color images and split to 50K training set and 10K test set. The classes in CIFAR-10 are natural objects such as airplane, automobile, bird and cat.

\subsection{Results}
We use two benchmark datasets, SVHN and CIFAR-10 to demonstrate the performance of our approach in SSL, by comparing with some state-of-art methods, such as Mean-Teacher \cite{tarvainen2017mean}, $\Pi$ model \cite{laine2016temporal}, Bad GAN \cite{dai2017good} and Triple GAN \cite{chongxuan2017triple}. As shown in \tref{aba:tbl1} and \tref{aba:tbl2}, our MLSN model outperforms all these methods with higher prediction accuracy with different proportions of labeled set in the whole dataset. 
% SVHN
\begin{table}
	\tbl{Error rate percentage on SVHN, averaged over 10 runs.}
	{\begin{tabular}{@{}cccc@{}}\toprule
			Model & 250 labels & 500 labels & 1000 labels \\
			\colrule
			Supervised Only \cite{tarvainen2017mean}& $42.65 \pm 2.68$ & $22.08 \pm 0.73$ & $14.46 \pm 0.71$ \\
			$\Pi$ model \cite{laine2016temporal}& $9.93 \pm 1.15$ & $6.65 \pm 0.53$ & $4.82 \pm 0.17$ \\
			Bad GAN \cite{dai2017good}& - & - & $4.27 \pm 0.03$ \\
			Triple GAN \cite{chongxuan2017triple}& - & - & $5.77 \pm 0.17$ \\
			MT & $40.92 \pm 1.13$ & $6.8 \pm 0.42$ & $5.12 \pm 0.08$ \\
			MT + Similarity (\textbf{ours}) & $15.2 \pm 0.82$ & $5.6 \pm 0.21$ & $4.63 \pm 0.04$ \\
			\botrule
		\end{tabular}
	}
	\label{aba:tbl1}
\end{table}

% CIFAR-10 
\begin{table}
	\tbl{Error rate percentage on CIFAR-10, averaged over 10 runs.}
	{\begin{tabular}{@{}cccc@{}}\toprule
			Model & 1000 labels & 2000 labels & 4000 labels \\
			\colrule
			Supervised Only \cite{tarvainen2017mean}& $46.43 \pm 1.21$ & $33.94 \pm 0.73$ & $20.66 \pm 0.57$ \\
			$\Pi$ model \cite{laine2016temporal}& $27.36 \pm 1.20$ & $18.02 \pm 0.60$ & $13.20 \pm 0.27$ \\
			Bad GAN \cite{dai2017good}& - & - & $14.41 \pm 0.03$ \\
			Triple GAN \cite{chongxuan2017triple}& - & - & $16.99 \pm 0.36$ \\
			MT & $22.90 \pm 0.93$ & $18.20 \pm 0.62$ & $13.30 \pm 0.30$ \\
			MT + Similarity (\textbf{ours}) & $18.87 \pm 0.92$ & $15.28 \pm 0.20$ & $11.20 \pm 0.17$ \\
			\botrule
		\end{tabular}
	}
	\label{aba:tbl2}
\end{table}

%\subsection{Weak label}
We can obtain a dateset 'weak-labeled data' if we do not know its corresponding label $(y_i,y_j)$ for an arbitrary instance pair $(x_i,x_j)$ but only know whether $x_i$ and $x_j$ belong to the same class. This kind of weak labels can exactly be fed into our well-designed similarity network $S$.  
%this label named after the weak label. In practice, high-quality labeled data is difficult to obtain, and weak-labeled data is easier to get and with higher quality usually.\par
%Noted that the input of similarity network is weak labeled data. More interesting is that we random select 1000 labeled data on the CIFAR-10 dataset, and convert other labeled data into weak labeled data. 

\begin{figure}
	\begin{center}
		\includegraphics[width=4.5in]{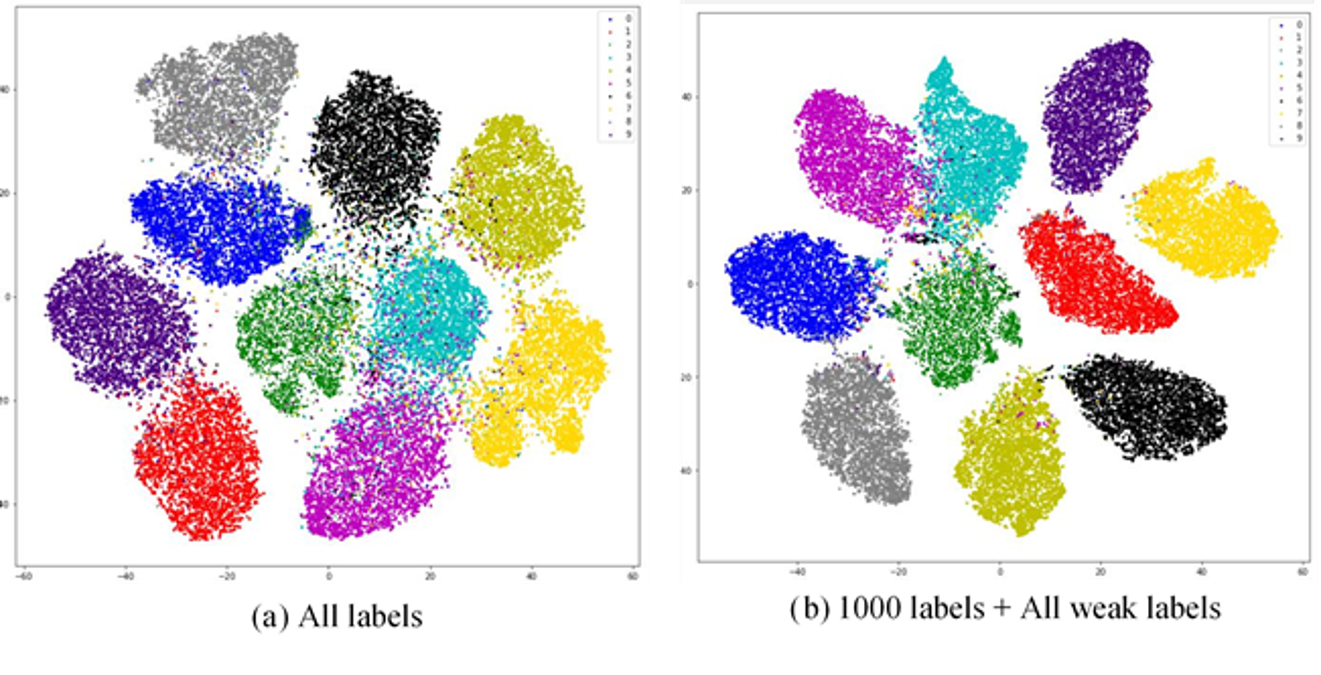}
	\end{center}
	\caption{(a,b) are the visualization of feature space using t-SNE \cite{maaten2008visualizing}. (a) utilized all labeled data,  while (b) only used 1000 labels and all weak labels}
	\label{aba:fig2}
\end{figure}

We compared the performance of different methods on weak-label dataset of 1000 instances randomly selected from CIFAR-10.
\fref{aba:fig2} plots the 2D projection of leanred features using T-SNE. It is obvious that, the representation learned by our model shows more clear and separable clustering. And the accuracy of our model is $94.7\%$ on the test set, slightly better than fully supervised learning. 
\section{Conclusion}
In this paper, we propose a novel method named Metric Learning by Similarity Network (MLSN) to learn a distance metric adaptively on different domains. By co-training with the classification network, the similarity network can catch more pairwise relationships to help increase the effectiveness of the classifier training. 

\bibliographystyle{ws-procs9x6} % for numbered citation & references
\bibliography{ws-pro-sample}

\end{document}